%% file: arxiv.tex
\PassOptionsToPackage{table}{xcolor}
\documentclass{mdraft_arxiv}

\input{math_commands.tex}

\usepackage[utf8]{inputenc}
\usepackage[T1]{fontenc}
\usepackage{hyperref}
\usepackage{url}
\usepackage{booktabs}
\usepackage{amsfonts}
\usepackage{nicefrac}
\usepackage{microtype}
\usepackage{amsmath,amssymb,mathtools}
\usepackage{array}
\usepackage{enumitem}
\usepackage{makecell}
\usepackage{caption}
\usepackage{graphicx}
\usepackage{wrapfig}
\usepackage{float}
\usepackage{placeins}
\usepackage{algorithm}
\usepackage{algpseudocode}
\usepackage{pgfplots}
\pgfplotsset{compat=1.18}

\title{Multi-Draft On-Policy Distillation\\for Diffusion Drafters}
\author{Petr Molodyk}
\author{Yuchen Zhu}
\author{Jaemoo Choi}
\author{\\Molei Tao}
\author{Yongxin Chen}
\affiliation{Georgia Institute of Technology}
\newcommand{\modellink}{\href{https://huggingface.co/FLAIR-Gatech/Qwen3-8B-MDraft-b16}{\raisebox{-0.3em}{\includegraphics[height=1.5em]{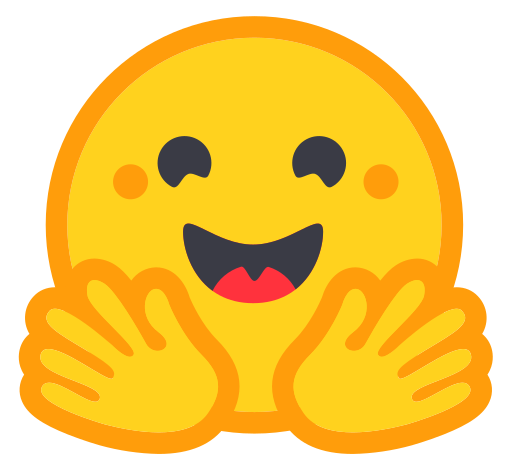}}\hspace{0.5em}\textsf{Checkpoint}}}

\abstract{Speculative decoding is used to reduce the inference cost of language models. One-step diffusion drafters propose a block of multiple tokens in parallel, reducing the sequential cost of speculative decoding. On-policy distillation trains these drafters using verifier feedback on their own proposals. However, parallel drafting makes this feedback difficult to use: the drafter predicts each token without observing the preceding tokens sampled within the same block, while the verifier conditions its prediction on those tokens. Different sampled prefixes therefore give different verifier targets for the same drafter prediction. We show that distilling against these targets separately can suppress tokens that are plausible after some prefixes but unlikely after others, harming draft acceptance. We explain the effects of this mismatch and introduce Multi-Draft On-Policy Distillation (MDraft), a new parallel drafter distillation method which samples several draft blocks from the same context and aggregates the verifier's full-vocabulary conditionals across their prefixes. We use all the drafts to collect information from the verifier, allowing support from alternative prefixes to influence the training target. The resulting algorithm improves the acceptance length of a parallel drafter beyond the level of pre-training, and alleviates the effects of the distributional mismatch. MDraft changes only training, retaining the drafter's architecture and inference procedure, allowing the resulting drafter to be used as-is in place of the pretrained checkpoint. Experiments on a suite of mathematical, coding, and conversational tasks show longer accepted drafts than prior distillation methods.}

\begin{document}
\maketitle

\section{Introduction}

\begin{figure}[t]
\centering
\includegraphics[width=0.88\linewidth]{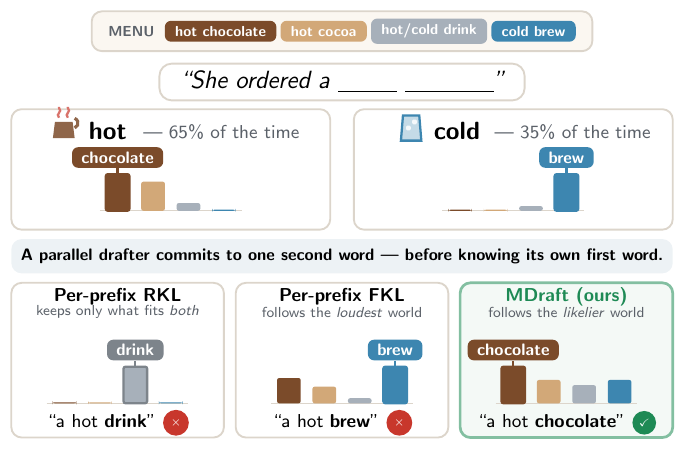}
\caption{A parallel drafter predicts the second token without observing its sampled first token. The verifier's preferred continuation depends on that first token; different training objectives therefore yield different second-token choices. The verifier's greedy continuation is ``hot chocolate.'' Bars are schematic; Table~\ref{tab:example} gives the numerical distributions.}
\label{fig:two-branch-example}
\end{figure}

Generating text with a large language model requires a sequence of expensive forward passes, one for each output token. Speculative decoding reduces this cost by allowing a smaller drafter model to propose several tokens and verifying them together with the target model \citep{leviathan2023speculative,chen2023speculative}. The verification procedure preserves the LLM's output distribution, while its efficiency depends on how many proposals are accepted relative to the cost of producing them. One-step diffusion drafters offer an attractive trade-off: they generate an entire block of tokens in parallel, avoiding the sequential forward passes required by an autoregressive drafter \citep{christopher2024specdiff,chen2026dflash}.

Training such drafters typically begins with supervision from target-generated continuations \citep{chen2026dflash,li2025eagle3}. Works in the LLM domain show that on-policy training helps further improve models by fine-tuning them on self-generated data \citep{gu2024minillm,agarwal2024gkd}, allowing the student model to be trained on the states that it actually encounters in inference. Recent works in parallel drafter training adopt this philosophy by generating proposal blocks with the drafter model itself and evaluating the verifier LLM on these drafts. Draft-OPD \citep{lei2026draftopd} shows that training on drafter-generated blocks can improve drafter performance even when off-policy training saturates.

Parallel drafters, however, carry a structural difference compared to autoregressive models: the tokens generated in different positions of the drafted block are only conditioned on the context, not on the previous drafted tokens, while the LLM, which is used to supply the targets for the training objective, processes this block autoregressively. Consequently, even plausible sampled prefixes can give conflicting verifier targets for the same drafter prediction: a token that is useful after one prefix may receive little support after another. Distilling these targets separately can therefore suppress tokens that would be useful in other drafts.

In this work, we show that this problem has real consequences for parallel drafter training and design a new on-policy drafter distillation objective, MDraft, that aggregates verifier scores across multiple drafts to build a reliable target for the drafter distribution. We conduct an empirical evaluation by fine-tuning parallel drafters and evaluating their acceptance rates on multiple benchmarks.

Our main contributions are:
\begin{itemize}[leftmargin=*,nosep]
\item \textbf{Identifying the target mismatch in parallel drafter distillation.} We show that parallel drafters introduce a unique challenge for on-policy distillation, and using verifier signals on drafter continuations as targets can favor tokens that are harmful to draft acceptance.
\item \textbf{Developing Multi-Draft On-Policy Distillation.} We introduce Multi-Draft On-Policy Distillation (MDraft), which combines verifier distributions on multiple drafts into a single target to address the mismatch.
\item \textbf{Improving on-policy distillation of parallel drafters.} MDraft achieves the best overall mean acceptance lengths among current on-policy distillation methods for parallel drafters, evaluated across a suite of mathematical, coding, and conversational benchmarks.
\end{itemize}

\section{Background}
\label{sec:background}

In this section, we introduce the main concepts and notation used in our paper. In Section \ref{sec:llm_sampling}, we discuss the mechanics of LLM sampling, and why speculative decoding is used. We then introduce parallel drafters for speculative decoding. We then explain how they are trained in Section \ref{sec:spec_training} and introduce on-policy training in Section \ref{sec:drafter_onpolicy}. Appendix~\ref{app:related-work} provides a broader discussion of related work.

\subsection{Language model sampling}
\label{sec:llm_sampling}
An autoregressive LLM generates each token from a probability distribution conditioned on the prompt and all previously generated tokens. Temperature controls the sharpness of this distribution, with greedy decoding selecting the most probable token. Standard decoding requires a separate model forward pass for each new token, making generation sequential.

\textbf{Speculative decoding.}
To reduce the number of expensive LLM passes, speculative decoding uses a cheaper drafter model to propose several tokens. In the simplest setup, the drafter is a smaller autoregressive language model that generates several tokens sequentially. The target LLM evaluates all these proposed tokens in one forward pass, using causal attention to compute the appropriate next-token distribution at each position. These tokens are either accepted or rejected, based on a special rule that ensures that the resulting sequence distribution matches that of the original LLM \citep{leviathan2023speculative,chen2023speculative}. Speedups arise when the cost of drafting and joint verification is lower than that of generating the committed tokens directly with the large model. For a fixed drafter and target architecture, the speedup becomes a function of the average \textit{acceptance length}: the average number of proposed tokens that the verifier model accepts in one pass.

 \textbf{Parallel drafting.} To further reduce the cost of speculative decoding, a \textit{parallel drafter} can be used. A parallel drafter predicts multiple future tokens at once. This removes the sequential passes of an autoregressive drafter, making longer proposals inexpensive to generate. The verifier can then check the block using the same speculative decoding procedure.

\textbf{Notation and conventions.} When discussing parallel drafting, we will denote the length of the drafted block as $L$. The parallel drafter model is denoted $q$, and for the categorical distribution it induces over the $\ell$-th token of the block we write $q_{\ell}(\cdot\mid c)$, where $c$ is the context that includes the prompt and previously accepted tokens. The current draft is denoted as $x$. The LLM model is $p$, also called the verifier, and the distribution is autoregressive: $p_\ell(\cdot\mid c, x_{<\ell})$. The vocabulary is denoted $\mathcal V$.
\subsection{Speculative drafter training}
\label{sec:spec_training}
Autoregressive drafters can be trained explicitly to match the target LLM. These methods increase draft acceptance lengths by aligning the distributions of the drafter model with the target LLM. Token-level supervision minimizes cross-entropy, $\mathcal L_{\mathrm{CE}}=-\sum_\ell\log q(x_\ell\mid c,x_{<\ell})$, on reference continuations $x$. Distributional distillation instead matches the target LLM's full next-token predictions: for a training prefix $h=(c,x_{<\ell})$, a common objective is $\mathrm{KL}(p(\cdot\mid h)\|q(\cdot\mid h))$. DistillSpec studies how the choice of divergence and training sequences affects acceptance, including sequences generated by the drafter itself \citep{zhou2024distillspec}.

\textbf{Parallel drafter training.} Similar loss functions can be used to train parallel drafters: DFlash is trained on responses generated by the verifier LLM with a cross-entropy loss: $\mathcal L_{\mathrm{DFlash}}=-\sum_{\ell=1}^{L}w_\ell\log q_\ell(x_\ell\mid c)$, where $x_\ell$ is the reference token generated by the LLM and the per-position weights are $w_\ell=\exp(- (\ell-1)/\gamma)$ for decay scale $\gamma>0$. These weights emphasize the importance of early predictions because early rejection prevents later proposals from being used \citep{chen2026dflash}. DSpark tries to re-introduce dependency between tokens in the draft by augmenting the parallel backbone with a lightweight sequential head and using a confidence head to schedule the verification length \citep{cheng2026dspark}. It combines the same position-weighted cross-entropy used in DFlash with distribution matching and confidence supervision: $\mathcal L_{\mathrm{DSpark}}=0.1\mathcal L_{\mathrm{DFlash}}+0.9\mathcal L_{\mathrm{TV}}+\mathcal L_{\mathrm{conf}}$. Here, $\mathcal L_{\mathrm{TV}}=\sum_\ell w_\ell\lVert q_\ell-p_\ell\rVert_1$ is a surrogate for the expected token acceptance, with drafter $q_\ell$ and LLM $p_\ell$ evaluated on LLM-generated prefixes. The confidence loss is $\mathcal L_{\mathrm{conf}}=\sum_\ell w_\ell\operatorname{BCE}(s_\ell,a_\ell^\star)$, where $s_\ell$ is the predicted acceptance probability, $a_\ell^\star=1-\tfrac12\lVert q_\ell-p_\ell\rVert_1$ is its soft target, and $\operatorname{BCE}$ is the binary cross-entropy loss.

\subsection{On-policy training}
\label{sec:drafter_onpolicy}
Drafter training is usually performed by computing drafter distributions over a corpus of data generated by the verifier LLM. However, inspired by the success of on-policy knowledge distillation in LLMs \citep{agarwal2024gkd}, recent works develop methods for on-policy drafter training.

LLM on-policy distillation trains an autoregressive student LLM on its own continuations, using teacher feedback on sequences generated by the student. For a prompt $c$ and previously generated tokens $y$, let $p(\cdot\mid c)$ and $q(\cdot\mid c)$ denote teacher and student sequence distributions, and $p_\ell,q_{\ell}$ their next-token distributions given $(c,y_{<\ell})$. The fully on-policy variant of generalized knowledge distillation (GKD) uses $\mathcal L_{\mathrm{GKD}}=\mathbb E_{y\sim q}[|y|^{-1}\sum_{\ell=1}^{|y|}\mathrm{KL}(q_{\ell} \| p_\ell)]$, averaged over prompts \citep{agarwal2024gkd}. MiniLLM instead targets sequence-level reverse KL, $\mathcal L_{\mathrm{MiniLLM}}=\mathbb E_{y\sim q}[\log q(y\mid c)-\log p(y\mid c)]$, again averaged over prompts \citep{gu2024minillm}. Both approaches obtain supervision on student-generated states, an idea subsequently adapted to speculative drafters.

\textbf{On-policy drafter training.} A drafter cannot by itself generate full sequences. Hence, on-policy training algorithms for drafters use the following protocol: first, a corpus of prompts is selected, and responses are generated by the verifier LLM. Then, a random position in the response is sampled and a proposal is generated at that position using the drafter model. The verifier LLM's distribution is then computed on this drafted block, and a distribution matching loss is calculated. 

Draft-OPD combines two losses: $\mathcal L_{\mathrm{Draft\text{-}OPD}}=\tfrac12(\mathcal L_{\mathrm{acc}}+\mathcal L_{\mathrm{rej}})$ \citep{lei2026draftopd}. Writing $\mathcal A$ and $\mathcal R$ for accepted and rejected draft positions, $\mathcal L_{\mathrm{acc}}=|\mathcal A|^{-1}\sum_{\ell\in\mathcal A}\mathrm{KL}(p_\ell\|q_\ell)$ matches the verifier on accepted states, while $\mathcal L_{\mathrm{rej}}=\sum_{\ell\in\mathcal R}u_\ell\mathrm{KL}(q_\ell\|p_\ell)$ suppresses unsupported draft predictions. The authors argue that using a combination of forward-KL and reverse-KL for the accepted and rejected tokens is useful, exploiting the mode-covering and mode-seeking properties of these divergences. The weights $u_\ell$ decay exponentially with the position within the block. AdaFlash uses $\mathcal L_{\mathrm{AdaFlash}}=\lambda\mathcal L_{\mathrm{hard}}+(1-\lambda)\mathcal L_{\mathrm{RKL}}^{\mathrm{clip}}$ \citep{qian2026adaflash}. Here, $\mathcal L_{\mathrm{hard}}=-L^{-1}\sum_\ell\log q_\ell(y_\ell^\star)$, with $y_\ell^\star=\arg\max_y p_\ell(y)$, encourages agreement with the verifier's preferred token. The distributional term is $\mathcal L_{\mathrm{RKL}}^{\mathrm{clip}}=L^{-1}\sum_{\ell,y}\min\{q_\ell(y)\log[q_\ell(y)/p_\ell(y)],\delta\}$: clipping each vocabulary-entry contribution at $\delta>0$ limits outlier gradients. The hard-loss component encourages agreement with the verifier's argmax, while the reverse-KL component penalizes excess draft probability on tokens that receive little verifier support.

\section{MDraft: Multi-Draft On-Policy Distillation}
\label{sec:method}

\subsection{On-policy objective for drafting}

A direct adaptation of LLM distillation objectives to parallel drafters is to sample a block from the current drafter, evaluate the verifier on that block, and minimize a token-level divergence such as forward or reverse KL. Writing $p_{\ell}=p_\ell(\cdot\mid c,x_{<\ell})$, the per-position reverse KL objective is $\mathbb E_{x\sim q(\cdot\mid c)}[\mathrm{KL}(q_\ell\| p_{\ell})]$.

However, the distributions being compared have different conditioning information. In autoregressive LLM distillation, both student and teacher predict the next token given the same generated prefix. A parallel drafter instead computes all $L$ distributions $q_\ell(\cdot\mid c)$ in one forward pass and samples tokens at all positions independently, giving the factorized block distribution $q(x_{1:L}\mid c)=\prod_{i=1}^{L}q_i(x_i\mid c)$. Its sampled prefixes therefore follow $q_{<\ell}(x_{<\ell}\mid c)=\prod_{i<\ell}q_i(x_i\mid c)$. The prediction $q_\ell(\cdot\mid c)$ is shared across all such prefixes, whereas the verifier conditions on the particular realization $x_{<\ell}$. The drafter must therefore fit one distribution to several potentially incompatible verifier conditionals. Even individually plausible draft tokens can form an incoherent prefix, making the following tokens unlikely under autoregressive law. Therefore, the verifier's feedback is correct for that prefix, but need not describe which tokens are useful at that position across other possible drafts.

\textbf{A per-prefix divergence penalizes both tokens that are completely unsuitable at a position and tokens that are useful there but incompatible with the particular prefix sampled in this draft.} Because the parallel drafter cannot condition its prediction on that prefix, suppressing the latter tokens also reduces their probability in drafts where they would fit. On-policy sampling exposes this conflict directly.

\textbf{Distribution mismatch in action.} To illustrate the problem, consider the example in Figure~\ref{fig:two-branch-example}, assuming every word is represented by a token. Take the sentence ``She ordered a,'' followed by either \emph{hot} or \emph{cold}. In our illustrative example, \emph{hot} is more likely, and the verifier then prefers \emph{chocolate} while also supporting \emph{cocoa}. After \emph{cold}, it strongly prefers the word \emph{brew}. The generic word \emph{drink} has support after either prefix but is preferred after neither. A parallel drafter predicts both positions together: its second-word distribution cannot change after the first word is sampled. Thus, \emph{chocolate} can receive a strong negative training signal after a sampled \emph{cold}, even though it is the preferred continuation after \emph{hot}. Simultaneously, \emph{brew} will receive a strong positive signal, since it is the preferred token after \emph{cold}, even though the verifier prefers \emph{hot} as the first continuation token.

Per-token divergences, such as commonly used forward- and reverse-KL, fall into these traps: reverse KL strongly discounts tokens with low probability under one of the continuations, preferring the ``safety'' token \emph{drink}, and the forward KL will favor \emph{brew} in the second position, even though the verifier prefers \emph{hot} in the first. Figure~\ref{fig:two-branch-example} illustrates these choices. We provide the exact calculation in Table~\ref{tab:example} and the distribution-level explanation in Section~\ref{sec:target-analysis}.

\subsection{The MDraft objective}
\label{sec:mdraft-objective}

\textbf{Verifying tokens across multiple drafts.} A single draft only tells us how the verifier evaluates a token after one particular prefix. To learn whether a low score is specific to that prefix, MDraft samples $K$ independent draft blocks from the same context and evaluates all of them with the verifier. At each position, these blocks provide different sampled prefixes under which the verifier can assess the same candidate token. For example, a token penalized after a draft containing \emph{cold} may receive strong support after another containing \emph{hot}. We collect this evidence by averaging the verifier's token probabilities across the $K$ drafts, then use it to inform each draft's training target.

\textbf{Prefix-averaged token distribution.} Concretely, one drafter forward pass on context $c$ produces the position-wise distributions used to sample $x_{1:L}^{(k)}\sim q(\cdot\mid c)$ independently for $k=1,\ldots,K$. Evaluating the verifier on draft $k$ gives $p_\ell^{(k)}(y)=p(y\mid c,x_{<\ell}^{(k)})$, the probability of candidate token $y$ at position $\ell$ given the $k$-th draft's prefix. We average these distributions over the $K$ prefixes as $\widehat p_\ell(y)=K^{-1}\sum_{k=1}^{K}p_\ell^{(k)}(y)$.

The average measures how much support a token receives across the sampled prefixes. A token with low probability under one prefix can get support from others, raising the average. Conversely, if every sampled prefix assigns it a low probability of at most $\varepsilon$, its average probability is also at most $\varepsilon$. \textit{Thus, averaging can preserve useful alternatives without raising the scores of tokens that are implausible across all sampled drafts}.

\textbf{Distillation target.} We use the average $\widehat p_\ell$ to bring this evidence from other drafts into each proposal's distillation target. The average captures support across prefixes, while the individual conditional $p_\ell^{(k)}$ retains information about compatibility with the sampled prefix. We blend the two so that support from alternative continuations can temper a low score under one prefix without discarding prefix-specific information entirely. The coefficient $\alpha\in[0,1]$ controls their balance through interpolation in log-probability space, making the target distribution:
\begin{equation}
 \widetilde r_\ell^{(k)}(y)=p_\ell^{(k)}(y)^{1-\alpha}\widehat p_\ell(y)^\alpha,
 \qquad r_\ell^{(k)}(y)=\frac{\widetilde r_\ell^{(k)}(y)}{Z_{\ell,k}},
 \quad Z_{\ell,k}=\sum_{z\in\mathcal V}\widetilde r_\ell^{(k)}(z).
 \label{eq:tilted-target}
\end{equation}
Here $Z_{\ell, k}$ is the normalizing constant.
At $\alpha=0$, $r_\ell^{(k)}=p_\ell^{(k)}$, recovering the naive per-prefix target; at $\alpha=1$, every proposal shares $r_\ell^{(k)}=\widehat p_\ell$.

We fit the drafter to this target using reverse KL. For a context $c$, the training objective is
\begin{equation}
 \mathcal L_{\mathrm{MDraft}}(c)=\frac{1}{L\cdot K}
 \sum_{k=1}^{K}\sum_{\ell=1}^{L}w_{\ell,k}\,
 \mathrm{KL}(q_\ell\|r_\ell^{(k)}).
 \label{eq:mdraft}
\end{equation}
We minimize its batch average, holding sampled prefixes, targets, and nonnegative position weights fixed during differentiation. Appendix~\ref{app:training-options} discusses weighting options $w_{\ell, k}$.

\par
\begingroup
\setlength{\fboxsep}{4pt}
\noindent\colorbox{juniperlight}{\parbox{\dimexpr\linewidth-2\fboxsep\relax}{MDraft retains the standard per-token reverse-KL distillation loss and changes the target it matches. Each target blends the verifier distribution for one draft with the average across multiple drafts, incorporating support from alternative prefixes into one distillation objective.}}
\par\endgroup

\textbf{Revisiting the two-branch example.} We can observe the benefit of MDraft's objective on the example from Figure~\ref{fig:two-branch-example}. Let the verifier and first-position drafter both assign probability $0.65$ to \emph{hot} and $0.35$ to \emph{cold}. Per-prefix RKL favors \emph{drink}, which has support on both branches, while FKL favors \emph{brew} because its average probability exceeds that of \emph{chocolate}. At $\alpha=0.9$, MDraft preserves \emph{chocolate}'s support from the more likely \emph{hot} branch while discounting \emph{brew} for being implausible on that branch. It therefore recovers the verifier's greedy second token and also achieves higher second-token acceptance at $T=1$ than either KL baseline. Table~\ref{tab:example} in Appendix~\ref{app:two-branch-example} provides the exact setup, optimal distributions, and acceptance values.

\subsection{Training algorithm and computational cost}

We present the training procedure of MDraft in algorithm~\ref{alg:mdraft}. We use one drafter forward pass per context and $K$ verifier evaluations, which can be batched into one pass of the verifier model. Draft blocks are sampled at training temperature $1$; verifier targets use full-support probabilities. Evaluation temperature is specified independently.

\begin{algorithm}[H]
\caption{MDraft: one training step}
\label{alg:mdraft}
\begingroup
\algrenewcommand\algorithmicrequire{\textbf{Input:}}
\algrenewcommand\algorithmicensure{\textbf{Output:}}
\begin{algorithmic}[1]
\Require drafter $q$, frozen verifier $p$, a batch of contexts; block length $L$, proposal count $K$, blending coefficient $\alpha$, learning rate $\eta$
\For{each context $c$ in the batch}
 \State Run the drafter once to predict distributions for $L$ next tokens.
 \State Sample $K$ independent draft blocks from these distributions at temperature $1$.
 \State Run the verifier on each block to score every position given its sampled prefix.
 \State Average the $K$ verifier distributions at each position to obtain $\widehat p_\ell$.
 \State Blend each verifier distribution with $\widehat p_\ell$ to form the target $r_\ell^{(k)}$ (\eqref{eq:tilted-target}).
 \State Compute position weights for each draft (Appendix~\ref{app:training-options}).
 \State Compute the weighted distillation loss (\eqref{eq:mdraft}).
\EndFor
\State Average the losses over the batch and update the drafter at learning rate $\eta$.
\Ensure Updated drafter $q$.
\end{algorithmic}
\endgroup
\end{algorithm}

\textbf{Acceptance-aware position weights.} D-PACE \citep{wu2026dpace} suggests using position weights that decay at a rate tied to acceptance probability. An advantage of on-policy training is that verifying the drafter's own proposals provides direct acceptance feedback: we can compute the exact conditional acceptance probability of each sampled token under the verification rule. Let $a_{m,k}=\min\{1,p_m^{(k)}(x_m^{(k)})/q_m(x_m^{(k)}\mid c)\}$ denote the probability of accepting the sampled token at position $m$ in proposal $k$, conditional on all preceding tokens being accepted. This motivates detached survival weights $w_{\ell,k}\propto\prod_{m<\ell}a_{m,k}$, normalized to sum to $L$ within each proposal. This provides a decreasing position weight that is grounded in real acceptance probability, rather than an arbitrary schedule or approximation. We use survival weighting for MDraft and isolate its effect in Table~\ref{tab:position-ablation}. Appendix~\ref{app:training-options} gives more details.

\section{Experiments}
\label{sec:experiments}

We first evaluate MDraft across benchmarks and draft positions, then study the number of proposals used during training. We also examine scaling to a larger verifier model. We conclude with a theoretical analysis of how target blending changes the distillation objective (Section~\ref{sec:target-analysis}).

\textbf{Metrics and evaluation.} We report mean \textit{acceptance length} $\tau$: the number of accepted draft tokens plus one correction (bonus) token per verifier pass, averaged first within each prompt and then across prompts. Higher values mean that each verifier pass produces more output tokens. All methods use the DFlash architecture, and the inference algorithm is unchanged. With the draft block length and inference configuration held fixed within each comparison, accepted draft length captures the efficiency gains from improved drafting; we therefore do not report speedup separately. We evaluate mathematics on GSM8K \citep{cobbe2021gsm8k}, MATH500 \citep{lightman2024verify}, and AIME25 \citep{maa2025aime}; coding on MBPP \citep{austin2021mbpp} and HumanEval \citep{chen2021humaneval}; and conversational generation on MT-Bench \citep{zheng2023mtbench}. The reported average across benchmarks is an unweighted mean of the six benchmark-level measurements. All methods within a comparison use the same evaluation prompts and decoding settings.

\subsection{Main results}
\label{sec:main-results}

\textbf{Models and methods.} We fine-tune a pretrained DFlash drafter with Qwen3-4B \citep{yang2025qwen3} as the verifier model. We implement the AdaFlash and Draft-OPD losses, per-prefix RKL and FKL, and MDraft. All fine-tuned variants start from the same checkpoint and use the same architecture and block length, with no method-specific inference changes. Table~\ref{tab:main-results} reports acceptance lengths at evaluation temperatures $T\in\{0,0.6,1\}$.

\textbf{Training data and optimization.} We sample prompts from Nemotron-Post-Training-Dataset-v2  \citep{nathawani2025nemotrondata} and sample 4 continuations from the verifier model for each prompt. We then truncate the sequence to a maximum length of 1024 tokens. We choose a blend of prompts from the math, code and chat sections of the dataset. Specifically, our set contains 667,192 teacher-generated sequences (61.52\% math, 19.24\% code, and 19.24\% chat). We sample one draft start position per sequence per epoch and train with AdamW for three epochs, using a global batch size of 32 and a peak learning rate $10^{-5}$. MDraft uses $\alpha=0.9$ and $K=8$.

\textbf{Evaluation.} We evaluate the drafter on six benchmarks at different sampling temperatures $T\in\{0,0.6,1\}$ with thinking mode disabled and a 2,048-token generation cap. For every fine-tuned method, we evaluate the final checkpoint. Appendix~\ref{app:experimental-details} gives more details.

\begin{table}[!htbp]
\centering
\caption{Distilling DFlash with a Qwen3-4B teacher. Each cell reports mean acceptance length $\tau$, including the bonus token (higher is better). Mean is the unweighted average across benchmarks. Bold and underlining mark the best and second-best results within each temperature and column.}
\label{tab:main-results}
\begingroup
\footnotesize
\setlength{\tabcolsep}{2pt}
\renewcommand{\arraystretch}{0.9}
\begin{tabular*}{\linewidth}{@{\extracolsep{\fill}}lccccccc@{}}
\toprule
& \multicolumn{3}{c}{\textsc{Math}} & \multicolumn{2}{c}{\textsc{Code}} & \textsc{Chat} & \\
\cmidrule(lr){2-4}\cmidrule(lr){5-6}\cmidrule(lr){7-7}
Method & GSM8K & MATH500 & AIME25 & MBPP & HumanEval & MT-Bench & Mean \\
\midrule
\multicolumn{8}{c}{Temperature $T=0$} \\
\midrule
DFlash & 6.469 & 8.023 & 7.379 & 6.023 & 6.581 & 4.119 & 6.432 \\
AdaFlash & \underline{6.793} & \textbf{8.339} & \textbf{7.709} & \underline{6.141} & \underline{6.713} & \textbf{4.329} & \underline{6.671} \\
Draft-OPD & 6.604 & 7.844 & 7.113 & 5.882 & 6.438 & 4.198 & 6.347 \\
RKL & 6.688 & 7.899 & 7.108 & 5.940 & 6.513 & 4.247 & 6.399 \\
MDraft & \textbf{6.907} & \underline{8.322} & \underline{7.640} & \textbf{6.187} & \textbf{6.740} & \underline{4.317} & \textbf{6.685} \\
\midrule
\multicolumn{8}{c}{Temperature $T=0.6$} \\
\midrule
DFlash & 6.168 & 7.365 & 6.044 & 5.599 & 6.214 & 3.964 & 5.892 \\
AdaFlash & 6.487 & \underline{7.747} & 6.413 & \underline{5.792} & \underline{6.469} & 4.101 & \underline{6.168} \\
Draft-OPD & 6.397 & 7.446 & \underline{6.448} & 5.616 & 6.273 & 4.065 & 6.041 \\
RKL & \underline{6.518} & 7.564 & 6.385 & 5.711 & 6.384 & \underline{4.109} & 6.112 \\
MDraft & \textbf{6.739} & \textbf{7.905} & \textbf{6.480} & \textbf{6.001} & \textbf{6.593} & \textbf{4.205} & \textbf{6.321} \\
\midrule
\multicolumn{8}{c}{Temperature $T=1$} \\
\midrule
DFlash & 5.655 & 6.489 & 4.953 & 5.145 & 5.684 & 3.720 & 5.274 \\
AdaFlash & 5.988 & 6.914 & 5.264 & 5.285 & 5.980 & 3.843 & 5.546 \\
Draft-OPD & 6.067 & 6.854 & 5.214 & 5.343 & 5.916 & 3.919 & 5.552 \\
RKL & \underline{6.300} & \underline{7.041} & \underline{5.374} & \underline{5.541} & \underline{6.039} & \underline{4.062} & \underline{5.726} \\
MDraft & \textbf{6.424} & \textbf{7.258} & \textbf{5.649} & \textbf{5.672} & \textbf{6.330} & \textbf{4.087} & \textbf{5.903} \\
\bottomrule
\end{tabular*}
\endgroup
\par
\end{table}

\textbf{Results.} MDraft achieves the best mean acceptance length across all temperature settings. At $T=0.6$ and $T=1$, MDraft achieves the highest acceptance length on all six benchmarks in Table~\ref{tab:main-results}. Its mean acceptance lengths are $6.321$ and $5.903$, compared with $6.168$ and $5.726$ for the strongest competing method at each temperature, respectively. The gains span mathematics, coding, and conversational generation.

\textbf{Comparison with standard distillation objectives.} We compare MDraft with per-prefix reverse KL (RKL) and forward KL (FKL). Table~\ref{tab:main-results} includes the RKL baseline; Appendix~\ref{app:fkl-baseline} reports FKL post-training dynamics. In our post-training setting, FKL fails to improve the pretrained DFlash checkpoint.

\textbf{Controlling for additional verifier evaluations.} To separate the effect of target blending from using more verifier evaluations, we also train per-prefix RKL with $K=1$ and $K=8$ ($\alpha=0$). Increasing the number of drafts does not give consistent improvement when the aggregation is not used, and both controls remain below MDraft (Figure~\ref{fig:rkl-proposal-count}).

\subsection{Where do acceptance gains arise?}
\label{sec:position-analysis}

\begin{figure}[!htbp]
\centering
\includegraphics[width=\linewidth]{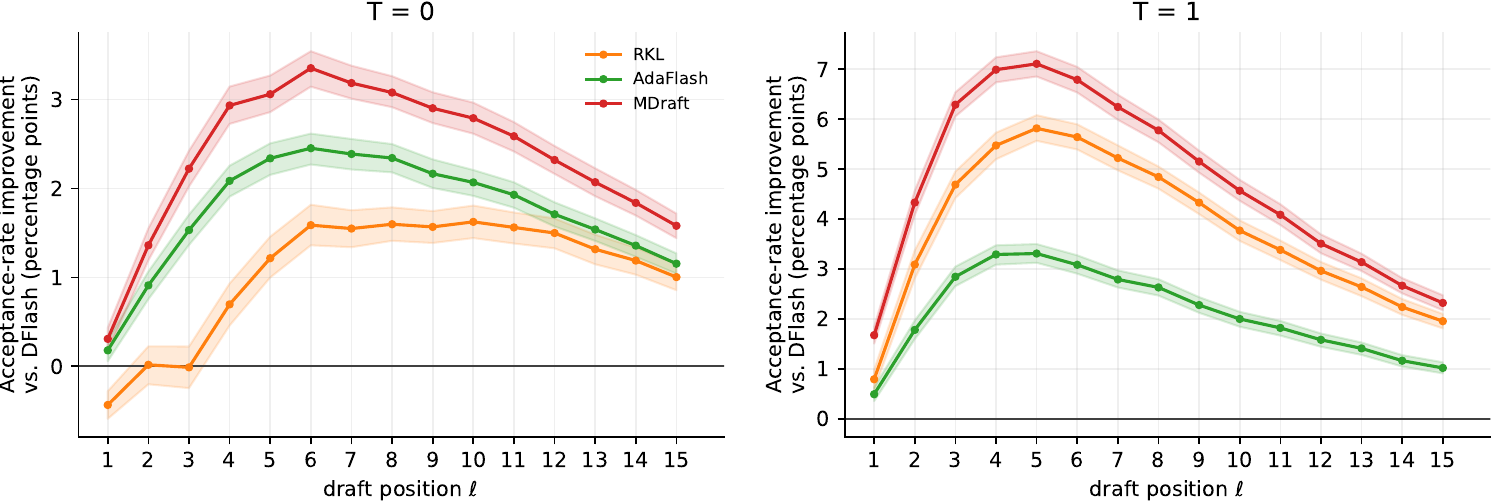}
\caption{Position-wise conditional acceptance gains over DFlash, in percentage points, at $T=0$ (left) and $T=1$ (right). Red: MDraft; green: AdaFlash; orange: RKL. The evaluation uses 1,656 prompts from GSM8K, MBPP and MT-Bench. Shaded bands denote 95\% confidence intervals. Positions index the 15 drafted tokens. MDraft's gains extend through the middle and later positions, with a larger advantage over AdaFlash at $T=1$.}
\label{fig:position-analysis}
\end{figure}

We use saved drafter checkpoints to examine how acceptance gains vary across the 15 drafted positions. Figure~\ref{fig:position-analysis} compares MDraft, AdaFlash and the RKL control, using DFlash as the reference and plotting the change in token acceptance rate conditioned on reaching each position.

MDraft improves conditional acceptance throughout the block. RKL yields larger gains under stochastic decoding than under greedy decoding, consistent with better distributional overlap without equally strong recovery of the verifier's argmax. AdaFlash shows the opposite trade-off relative to RKL: larger gains at $T=0$ but smaller gains at $T=1$. This pattern is consistent with its hard-target term favoring argmax agreement over matching the probabilities of alternative tokens. \textbf{MDraft improves acceptance in both greedy and stochastic decoding.} The gains extend through the middle and later positions, where verifier targets depend on more sampled draft tokens. Section~\ref{sec:target-analysis} examines how target aggregation can account for this behavior.

\subsection{Effects of proposal count}
\label{sec:training-dynamics}
\label{sec:ablations}

To establish how additional drafts improve the performance of MDraft, we use Qwen3-4B and the same pretrained DFlash checkpoint and training corpus as in the main experiment. We keep position weighting and all other training settings fixed across runs and vary the number of used drafts $K$.

\begin{figure}[!htbp]
\centering
\includegraphics[width=\linewidth]{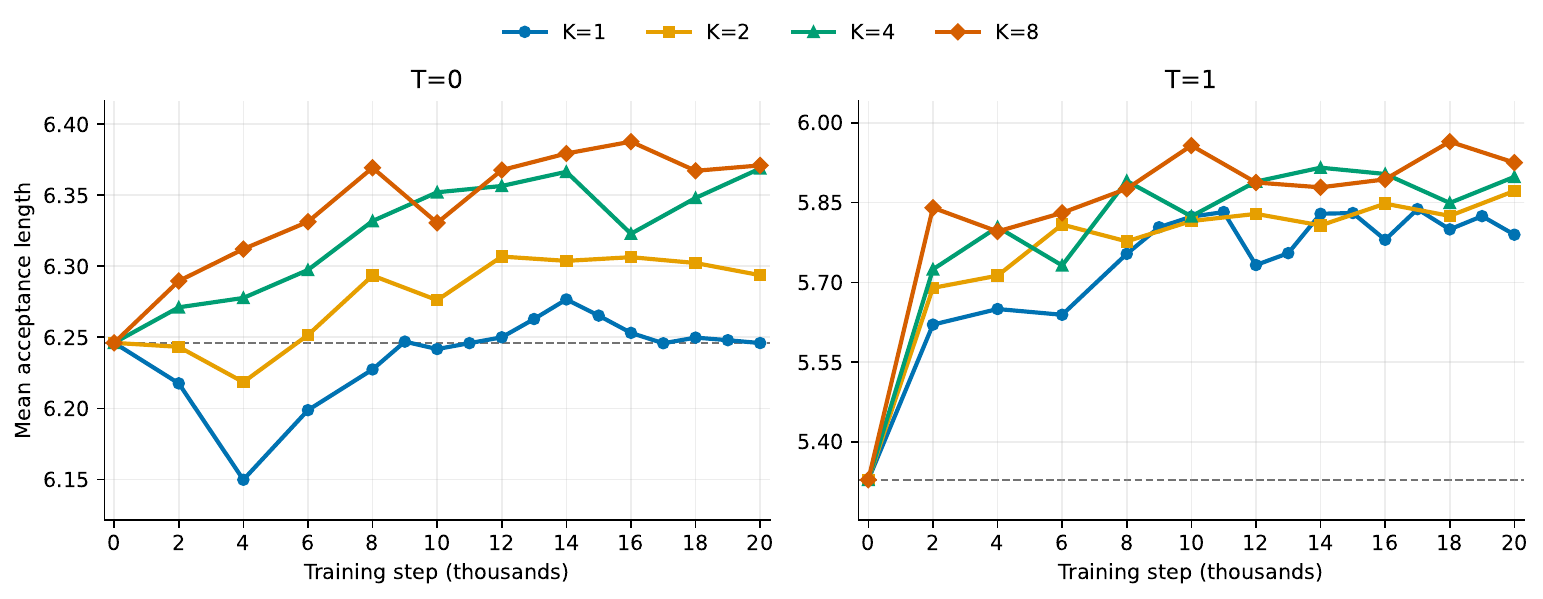}
\caption{We fix $\alpha=0.9$ and train DFlash for 20,000 iterations with varying numbers of parallel drafts $K=1,2,4,8$. We evaluate acceptance length on a subset of GSM8K and MBPP.}
\label{fig:training-dynamics}
\end{figure}

At $\alpha=0.9$, we vary $K\in\{1,2,4,8\}$ and track mean acceptance length throughout training (Figure~\ref{fig:training-dynamics}). All curves begin at the common pretrained checkpoint. The $K=1$ control reduces to per-prefix reverse KL. This comparison measures how additional proposals affect learning dynamics at a fixed number of training states.

Using multiple proposals yields a clear improvement over $K=1$, which finishes close to the initial checkpoint at $T=0$. At $T=1$, all settings improve acceptance, with the larger proposal counts reaching higher values.

\subsection{Scaling to a larger model}
\label{sec:scaling}

We compare DFlash, AdaFlash, and MDraft with Qwen3-4B and Qwen3-8B to examine how post-training gains change with verifier size. Training data are collected using a procedure similar to that of the main experiment (Appendix~\ref{app:scaling-data}). Table~\ref{tab:scaling} reports acceptance length on each benchmark at three evaluation temperatures, together with the relative improvement in mean acceptance length over the corresponding DFlash baseline. We observe that MDraft scales favorably to a larger model, with even larger gains in $T=1$ with the 8B model.

\begin{table}[!htbp]
\centering
\footnotesize
\caption{Post-training with Qwen3-4B and Qwen3-8B. Each cell reports mean acceptance length; MDraft uses $\alpha=0.9$ and $K=8$ for the 4B model and $\alpha=1.0$, $K=4$ for the 8B model. Bold marks the best result in each column within each model and temperature.}
\label{tab:scaling}
\begingroup
\setlength{\tabcolsep}{2pt}
\renewcommand{\arraystretch}{0.95}
\begin{tabular*}{\linewidth}{@{\extracolsep{\fill}}lccccccc@{}}
\toprule
Method & GSM8K & MATH500 & AIME25 & MBPP & HumanEval & MT-Bench & Mean \\
\midrule
\multicolumn{8}{c}{\textbf{Qwen3-4B}} \\
\midrule
\multicolumn{8}{c}{Temperature $T=0$} \\
DFlash & 6.469 & 8.023 & 7.379 & 6.023 & 6.581 & 4.119 & 6.432 \\
AdaFlash & 6.793 & \textbf{8.339} & \textbf{7.709} & 6.141 & 6.713 & \textbf{4.329} & 6.671 (+3.7\%) \\
MDraft & \textbf{6.907} & 8.322 & 7.640 & \textbf{6.187} & \textbf{6.740} & 4.317 & \textbf{6.685 (+3.9\%)} \\
\addlinespace[2pt]
\multicolumn{8}{c}{Temperature $T=0.6$} \\
DFlash & 6.168 & 7.365 & 6.044 & 5.599 & 6.214 & 3.964 & 5.892 \\
AdaFlash & 6.487 & 7.747 & 6.413 & 5.792 & 6.469 & 4.101 & 6.168 (+4.7\%) \\
MDraft & \textbf{6.739} & \textbf{7.905} & \textbf{6.480} & \textbf{6.001} & \textbf{6.593} & \textbf{4.205} & \textbf{6.321 (+7.3\%)} \\
\addlinespace[2pt]
\multicolumn{8}{c}{Temperature $T=1$} \\
DFlash & 5.655 & 6.489 & 4.953 & 5.145 & 5.684 & 3.720 & 5.274 \\
AdaFlash & 5.988 & 6.914 & 5.264 & 5.285 & 5.980 & 3.843 & 5.546 (+5.2\%) \\
MDraft & \textbf{6.424} & \textbf{7.258} & \textbf{5.649} & \textbf{5.672} & \textbf{6.330} & \textbf{4.087} & \textbf{5.903 (+11.9\%)} \\
\midrule
\multicolumn{8}{c}{\textbf{Qwen3-8B}} \\
\midrule
\multicolumn{8}{c}{Temperature $T=0$} \\
DFlash & 6.475 & 8.020 & 7.240 & 5.945 & 6.558 & 4.184 & 6.404 \\
AdaFlash & 6.784 & \textbf{8.411} & \textbf{7.653} & \textbf{6.122} & \textbf{6.715} & 4.243 & \textbf{6.655 (+3.9\%)} \\
MDraft & \textbf{6.831} & 8.316 & 7.528 & 6.114 & 6.706 & \textbf{4.292} & 6.631 (+3.5\%) \\
\addlinespace[2pt]
\multicolumn{8}{c}{Temperature $T=0.6$} \\
DFlash & 6.081 & 7.293 & 5.942 & 5.468 & 6.053 & 4.000 & 5.806 \\
AdaFlash & 6.452 & 7.818 & 6.479 & 5.752 & 6.314 & 4.050 & 6.144 (+5.8\%) \\
MDraft & \textbf{6.627} & \textbf{7.841} & \textbf{6.646} & \textbf{5.904} & \textbf{6.440} & \textbf{4.151} & \textbf{6.268 (+8.0\%)} \\
\addlinespace[2pt]
\multicolumn{8}{c}{Temperature $T=1$} \\
DFlash & 5.434 & 6.284 & 4.933 & 4.841 & 5.328 & 3.602 & 5.070 \\
AdaFlash & 5.971 & 6.835 & 5.516 & 5.207 & 5.740 & 3.864 & 5.522 (+8.9\%) \\
MDraft & \textbf{6.294} & \textbf{7.123} & \textbf{5.554} & \textbf{5.510} & \textbf{5.940} & \textbf{4.066} & \textbf{5.748 (+13.4\%)} \\
\bottomrule
\end{tabular*}
\endgroup
\end{table}

\subsection{How target blending changes the optimum}
\label{sec:target-analysis}
The experiments show that using an aggregated teacher in the distillation target improves drafting. To understand why, we ask which token distribution each objective would learn if it could be optimized exactly. We isolate one context and position, hold the draft-prefix distribution fixed, and use uniform position weights. We first analyze the population objective using the exact average verifier distribution, then account for the finite number of proposals used in practice. Write $p_x(y)$ for the verifier's probability of token $y$ after a sampled prefix $x$, and $\bar p(y)=\mathbb E_x[p_x(y)]$ for its average.

\textbf{What ordinary KL learns.} The parallel drafter must fit a single distribution $q$ to all of these prefix-dependent targets. In the optimum, forward KL averages their probabilities, while reverse KL averages their log probabilities \citep{genest1986pooling}:
\begin{align}
 q^\star_{\mathrm{FKL}}(y)&=\bar p(y),
 \label{eq:linpool}\\
 q^\star_{\mathrm{RKL}}(y)&\propto\exp\!\left(\mathbb E_x[\log p_x(y)]\right).
 \label{eq:logpool-bg}
\end{align}
The first solution preserves a token's average support, but can favor a confident minority branch, as with \emph{brew} in Table~\ref{tab:example}. The second strongly discounts a token if some prefixes give it very low probability. It can therefore favor a generic token such as \emph{drink} over a token that is preferred on one branch but incompatible with another.

\textbf{What MDraft changes.} To quantify this discount, define $J(y)=\log\bar p(y)-\mathbb E_x[\log p_x(y)]\geq0$. It is zero when the verifier assigns the same probability to $y$ under every prefix, and grows when substantial average support coexists with very low probabilities under some prefixes. MDraft's reverse-KL loss differs from the following expression only by a constant independent of $q$:
\begin{equation}
 \mathcal L_{\mathrm{MDraft}}(q)=\mathrm{KL}(q\|\bar p)+(1-\alpha)\sum_y q(y)J(y) + C
 \label{eq:mdraft-decomposition}
\end{equation}
The first term encourages the drafter to match average verifier support. The second charges a penalty for placing probability on tokens whose support varies across prefixes. Ordinary per-prefix reverse KL includes this penalty at full strength ($\alpha=0$). Blending with the average reduces its strength to $1-\alpha$, so incompatibility with one prefix has less influence on the shared drafter distribution.

\textbf{Consequence for token selection.} Minimizing this loss gives
\begin{equation}
 q_\alpha^\star(y)\propto\bar p(y)\exp\!\big(-(1-\alpha)J(y)\big).
 \label{eq:mdraft-optimum}
\end{equation}
Each token's average support is multiplied by a discount for prefix disagreement. At $\alpha=0$, this recovers the reverse-KL solution; at $\alpha=1$, the discount disappears and the solution equals the forward-KL optimum. An intermediate value can preserve useful branch-specific tokens while still penalizing inconsistent support. In the example illustrated in Figure~\ref{fig:two-branch-example}, it prevents \emph{chocolate} from being suppressed in favor of \emph{drink}, while discounting \emph{brew} enough to recover the verifier's greedy choice.

\textbf{Finite $K$ also retains a disagreement penalty.} The analysis above uses the exact average $\bar p$, which we cannot compute and therefore estimate during training with the sample average $\widehat p$. Although $\widehat p$ is unbiased, the loss uses its logarithm: concavity gives $\mathbb E[\log\widehat p(y)]\leq\log\bar p(y)$. Define this gap, which will depend on the sample size $K$, as $J_K(y)=\log\bar p(y)-\mathbb E[\log\widehat p(y)]$. In the expected training objective, the disagreement penalty becomes $(1-\alpha)J(y)+\alpha J_K(y)$ (Appendix~\ref{app:proposal-count}). Thus, finite $K$ leaves a residual penalty even at $\alpha=1$, qualitatively resembling a smaller blending coefficient. As $K$ grows, $J_K$ vanishes. At $\alpha=1$, the limiting loss is reverse KL to $\bar p$: it shares the forward-KL optimum, but generally not its gradient away from that optimum.
The implicit blending induced by finite $K$ depends on the data and can differ across tokens. The explicit $\alpha$ provides a common, continuous control over the contribution of the average target, without requiring a reduction in the sampling budget. The parameters are complementary: increasing $K$ improves estimation of the average, while $\alpha$ controls how strongly it enters the target. At finite $K$, however, a single $\alpha$ cannot generally compensate for the token-dependent bias $J_K(y)$. MDraft thus connects per-prefix RKL at $\alpha=0$ for any $K$ to the FKL optimum at $\alpha=1$ as $K\to\infty$, with a residual disagreement penalty between these limits. The post-training comparisons with RKL (Table~\ref{tab:main-results}) and FKL (Appendix~\ref{app:fkl-baseline}) support the usefulness of this intermediate regime.
\section{Conclusion}
In this work, we identify a unique challenge that affects on-policy training of parallel drafters, explain the mathematical mechanisms by which it arises and its practical consequences. Based on this analysis, we introduce MDraft: a new method for parallel drafter distillation that uses multiple drafts drawn from the same context to construct a reliable target distribution using the verifier model. We conduct experiments that demonstrate that our method improves the drafter beyond off-policy training and establish the role of our method in improving drafter acceptance lengths on several diverse benchmarks.

\beginappendix

\section{Related Work}
\label{app:related-work}

\textbf{Speculative decoding and drafter architectures.} Work following the original speculative decoding algorithms \citep{leviathan2023speculative,chen2023speculative} improves the trade-off between proposal quality and drafting cost. Medusa predicts future tokens with multiple heads, while Hydra introduces dependencies between those heads \citep{cai2024medusa,ankner2024hydra}. EAGLE drafts using target-model features; EAGLE-2 improves candidate-tree construction, and EAGLE-3 develops the training procedure and feature representation further \citep{li2024eagle,li2024eagle2,li2025eagle3}.

\textbf{Parallel and diffusion drafting.} Masked and block diffusion language models provide a framework for predicting several tokens together \citep{sahoo2024mdlm,arriola2025blockdiffusion}. Speculative Diffusion Decoding applies diffusion models to drafting, and DFlash uses target hidden states to produce a draft block in one step \citep{christopher2024specdiff,chen2026dflash}. DART also predicts future logits in a single pass from target hidden states, using position-weighted forward KL for training and consistency-aware tree pruning at inference \citep{liu2026dart}. Parallel drafting reduces sequential computation but makes dependencies between proposed tokens harder to represent. DSpark combines semi-autoregressive generation with confidence-scheduled verification, while Domino and TreeFlash introduce mechanisms for causal modeling within parallel drafting \citep{cheng2026dspark,huang2026domino,rheinboldt2026treeflash}. DBLAST and LiLiCorr address dependencies between proposed tokens to improve their joint consistency \citep{karimi2026dblast,rusanovsky2026lilicorr}. xPress addresses the same lack of within-block dependencies through a lightweight causal refiner \citep{wang2026xpress}. PRESTO identifies the mismatch between prefix-blind diffusion marginals and prefix-dependent verification, and uses prefix-aligned scoring for tree search \citep{wang2026presto}. MDraft instead changes the distillation targets while retaining the drafter architecture and inference procedure. Related inference methods explore block-diffusion draft trees, target-aware prefix selection, and extensions of parallel drafting \citep{ringel2026ddtree,wang2026taps,zhang2026d2sd}.

\textbf{Acceptance-aware training.} DistillSpec studies how knowledge distillation and the choice of divergence affect speculative decoding \citep{zhou2024distillspec}. More recent objectives account explicitly for the fact that verification retains a prefix rather than independently accepting positions. DFlash applies a fixed positional decay, while GRIFFIN, left-to-right diffusion training, D-PACE, and Spec-AUF modify the weighting or support of token-level supervision to emphasize errors that limit the accepted prefix \citep{chen2026dflash,hu2025griffin,whalen2026l2r,wu2026dpace,yang2026specauf}. Other approaches change the objective itself. LK Losses uses negative log acceptance and a scheduled KL--TV combination, and Variational Speculative Decoding formulates training in terms of sequence acceptance \citep{samarin2026lklosses,zou2026vsd}. SpecDiff-2 optimizes a streak-based objective on verifier-generated trajectories \citep{sandler2025specdiff2}.

\textbf{On-policy distillation.} ImitKD addresses exposure bias by querying the teacher on student-generated prefixes \citep{lin2020imitkd}. MiniLLM and generalized knowledge distillation study learning from student-generated sequences and the role of divergence choice in language-model compression \citep{gu2024minillm,agarwal2024gkd}. DistiLLM studies skew KL objectives and adaptive off-policy reuse of student-generated outputs \citep{ko2024distillm}; its teacher--student distribution mixing differs from our aggregation of teacher predictions across sampled prefixes. Online Speculative Decoding adapts draft models to observed queries through knowledge distillation \citep{liu2023online}, addressing deployment-time distribution shifts. For speculative draft models, Draft-OPD constructs supervision from verification outcomes, combining forward KL on accepted positions with reverse KL on rejected positions \citep{lei2026draftopd}. AdaFlash performs on-policy adaptation during deployment using a hard verifier target together with clipped reverse KL \citep{qian2026adaflash}. Our geometric-pooling analysis isolates the unmodified reverse-KL term; it does not characterize the full clipped or verification-dependent losses of these methods. JetSpec's comparison of forward and reverse KL for parallel draft heads also highlights the importance of objective choice in this setting \citep{hu2026jetspec}.

\textbf{Multimodality in parallel generation.} Independent output positions also create difficulties in representing multiple coherent continuations in non-autoregressive machine translation \citep{gu2018nat}. Distillation can reduce the complexity of the training distribution and make it easier for a parallel model to fit \citep{zhou2020natkd}. MDraft addresses the related issue of conflicting prefix-dependent supervision during on-policy drafter training, while the verifier preserves the final output distribution.

\section{Analysis of Target Blending}
\label{app:target-blending}

Consider reverse KL with fixed, proposal-independent weights. Fix a context $c$ and position $\ell$, and suppress them in the notation: write $q=q_\ell(\cdot\mid c)$, $p^{(k)}=p_\ell^{(k)}$, $r^{(k)}=r_\ell^{(k)}$, $\widehat p=\widehat p_\ell$, and $Z_k=Z_{\ell,k}$. The normalized-target loss is related to a weighted combination of reverse-KL losses by
\begin{align}
 R_{\alpha,K}(q)
 &=(1-\alpha)\frac1K\sum_{k=1}^{K}\mathrm{KL}(q\|p^{(k)})
   +\alpha\,\mathrm{KL}(q\|\widehat p),\nonumber\\
 \frac1K\sum_{k=1}^{K}\mathrm{KL}(q\|r^{(k)})
 &=R_{\alpha,K}(q)+\frac1K\sum_{k=1}^{K}\log Z_k.
 \label{eq:rkl-tilt-equivalence}
\end{align}
The normalization term is detached, so these objectives have identical drafter gradients. We analyze the equivalent objective $R_{\alpha,K}$. Hold the draft-prefix law $q_{<\ell}$ fixed. For a prefix $x\sim q_{<\ell}$, write $p_x(y)=p(y\mid c,x)$ and $\bar p(y)=\mathbb E_x[p_x(y)]$; thus $p^{(k)}=p_{x_{<\ell}^{(k)}}$. Assuming positive probabilities, define
\begin{equation}
  J(y)=\log\bar p(y)-\mathbb E_x[\log p_x(y)]\geq 0.
  \label{eq:prefix-disagreement}
\end{equation}
Adding and subtracting $\log\bar p(y)$ inside the reverse-KL objective gives
\begin{equation}
  \mathbb E_x\big[\mathrm{KL}(q\|p_x)\big]
  =\mathrm{KL}(q\|\bar p)+\sum_y q(y)J(y).
  \label{eq:rkl-decomposition}
\end{equation}
Thus, per-prefix reverse KL combines matching the marginal target with a penalty determined by how much each token's verifier probability varies across prefixes. The penalty is zero for a token whose probability is identical under all prefixes. It can be large for a token that is strongly supported under some prefixes but assigned near-zero probability under others. For a fixed prefix law, the drafter can lower this term by shifting probability away from tokens with large $J(y)$, even when they have substantial marginal support.

In the population limit, substituting the decomposition above into the two-term loss yields \eqref{eq:mdraft-decomposition}. Minimizing over a normalized probability vector gives \eqref{eq:mdraft-optimum}.

\subsection{Two-branch example}
\label{app:two-branch-example}
For the example in Figure~\ref{fig:two-branch-example}, fix the first-position drafter to match the verifier: $q_1(\text{hot})=0.65$ and $q_1(\text{cold})=0.35$. Table~\ref{tab:example} gives the two verifier conditionals and the optimal shared second-position distribution $q$ under each objective. We use uniform position weights and the exact branch average (the population limit for MDraft). The solutions follow from \eqref{eq:linpool}, \eqref{eq:logpool-bg}, and \eqref{eq:mdraft-optimum}.

At $T=0$, both models choose tokens greedily, so the first token is \emph{hot} and $A_0$ is one exactly when the drafter's second-token argmax is \emph{chocolate}. At $T=1$, the first token is always accepted because its draft and verifier distributions agree. The expected second-token acceptance is therefore
\[
 A_1(q)=0.65\sum_y\min\{q(y),p(y\mid\text{hot})\}
       +0.35\sum_y\min\{q(y),p(y\mid\text{cold})\}.
\]
MDraft at $\alpha=0.9$ selects \emph{chocolate} greedily and improves this overlap relative to both KL optima. The example demonstrates that an intermediate pooling objective can improve acceptance under both greedy and stochastic decoding.

\begin{table}[H]
\centering
\small
\caption{Numerical two-branch example for Figure~\ref{fig:two-branch-example}, with $\delta=10^{-9}$ and fixed branch probabilities $0.65/0.35$. Top: normalized verifier conditionals. Bottom: optimal second-position draft probabilities, rounded to three decimals, and second-token acceptance $A_T$ at evaluation temperature $T$. Bold probabilities mark each drafter's greedy choice; bold acceptance values mark the best of the three objectives.}
\label{tab:example}
\begingroup
\setlength{\tabcolsep}{4pt}
\begin{tabular*}{\linewidth}{@{\extracolsep{\fill}}lcccccc@{}}
\toprule
Verifier conditional & \emph{chocolate} & \emph{cocoa} & \emph{drink} & \emph{brew} & & \\
\midrule
$p(\cdot\mid\text{hot})$, $65\%$ & $0.42$ & $0.31$ & $0.27-\delta$ & $\delta$ & & \\
$p(\cdot\mid\text{cold})$, $35\%$ & $0.02$ & $0.01$ & $0.10$ & $0.87$ & & \\
\midrule
Objective & \emph{chocolate} & \emph{cocoa} & \emph{drink} & \emph{brew} & $A_0$ & $A_1$ \\
\midrule
Per-prefix RKL & 0.338 & 0.217 & \textbf{0.445} & 0.000 & 0 & 0.582 \\
Per-prefix FKL & 0.280 & 0.205 & 0.210 & \textbf{0.305} & 0 & 0.604 \\
MDraft ($\alpha=0.9$) & \textbf{0.350} & 0.253 & 0.278 & 0.119 & \textbf{1} & \textbf{0.655} \\
\bottomrule
\end{tabular*}
\endgroup
\end{table}

\section{The Role of the Number of Proposals}
\label{app:proposal-count}
Using the fixed-context, fixed-position notation of Appendix~\ref{app:target-blending}, the empirical average $\widehat p(y)=K^{-1}\sum_{k=1}^{K}p^{(k)}(y)$ is an unbiased estimator of $\bar p(y)$, but its logarithm is not. Define
$J_K(y)=\log\bar p(y)-\mathbb E[\log\widehat p(y)]$,
where the expectation is over the $K$ independent proposals. For the reverse-KL surrogate in \eqref{eq:rkl-tilt-equivalence},
\begin{equation}
  \mathbb E[R_{\alpha,K}(q)]
  =\mathrm{KL}(q\|\bar p)
  +\sum_y q(y)\big[(1-\alpha)J(y)+\alpha J_K(y)\big].
  \label{eq:mdraft-finite-k}
\end{equation}
By concavity of the logarithm, $0\leq J_K(y)\leq J(y)$. With $K=1$, $J_K=J$, and the reverse-KL branch reduces to per-prefix reverse KL for every $\alpha$. More generally, $K=1$ makes $r^{(1)}=p^{(1)}$, so $\alpha$ has no effect for any choice of divergence. As $K$ increases, the residual penalty is nonincreasing and converges to zero for a fixed finite vocabulary and block length with positive verifier probabilities. Consequently, $K$ controls both training cost and how closely the empirical objective approaches the population interpolation. At the first block position, all proposals share the same verifier conditional, so marginalization has no effect. The same holds at any position where the sampled conditionals agree. These target-construction properties hold for other divergences as well; the disagreement decomposition above is specific to reverse KL.

\phantomsection\label{app:divergences}
\textbf{Divergence choices.} MDraft fits the drafter distribution $q$ to each normalized tilted target $r^{(k)}$ by minimizing a divergence between them. Our implementation supports forward and reverse-KL, as well as Jensen-Shannon divergence. In the paper and the experiments, we always use reverse KL as the divergence in MDraft. We leave the analysis of the effects of these divergences to future work.

\section{Position Weighting}
\label{app:training-options}

The weights $w_{\ell,k}$ in \eqref{eq:mdraft} control the contribution of position $\ell\in\{1,\ldots,L\}$ in proposal $k$. We distinguish three choices below. All weights are held fixed when differentiating the loss; none changes the unweighted average $\widehat p_\ell$ used to construct the distillation targets.

\textbf{Uniform weights.} Setting $w_{\ell,k}=1$ gives every position in every proposal equal weight. The objective is then a simple average of the position-wise divergences, regardless of how likely verification is to reach each position.

\textbf{DFlash weights.} DFlash \citep{chen2026dflash} uses a fixed exponential decay with position. In our notation, this schedule is
\begin{equation}
  w_{\ell,k}=\exp\!\left(-\frac{\ell-1}{\gamma}\right),
  \qquad \gamma>0.
\end{equation}
The first position has weight one, and later positions receive progressively less weight. Smaller $\gamma$ concentrates supervision near the start of the block; as $\gamma\to\infty$, the weights approach uniform weighting. The schedule is shared across contexts and proposals: it favors earlier positions without using their actual acceptance probabilities.

\textbf{Survival weights.} Instead of prescribing a fixed decay, survival weighting adapts to each proposal. For a sampled token $x_m^{(k)}$, define its conditional acceptance probability under speculative rejection sampling as
\begin{equation}
  a_{m,k}=\min\!\left\{1,
  \frac{p_m^{(k)}(x_m^{(k)})}{q_m(x_m^{(k)}\mid c)}\right\}.
\end{equation}
Here $p_m^{(k)}=p(\cdot\mid c,x_{<m}^{(k)})$, and acceptance is conditional on all preceding draft tokens having been accepted. Thus, the probability that verification reaches position $\ell$ in this sampled proposal is
\begin{equation}
  \rho_{\ell,k}=\prod_{m=1}^{\ell-1}a_{m,k},
  \qquad
  w_{\ell,k}=\operatorname{sg}\!\left[
  \frac{L\rho_{\ell,k}}{\sum_{j=1}^{L}\rho_{j,k}}\right].
\end{equation}
The empty product is one, and $\operatorname{sg}$ denotes stop-gradient. Normalization gives $\sum_{\ell=1}^L w_{\ell,k}=L$, preserving the total weight of uniform supervision. The product includes only preceding positions: a token is still supervised if verification reaches it, even when that token itself is likely to be rejected. Positions after a likely rejection receive less weight. With fully greedy decoding, exact survival instead reduces to an indicator that all preceding draft tokens match the verifier's greedy choices.

\textbf{Position-weight comparison.} Table~\ref{tab:position-ablation} compares uniform and survival weighting for AdaFlash and MDraft in the Qwen3-4B main-experiment setup. MDraft uses $\alpha=0.9$, $K=8$, reverse KL; AdaFlash retains its hard-target and clipped reverse-KL terms. Survival weights are normalized to preserve the uniform loss scale. MDraft consistently performs better with survival weights compared to the uniform-weight setup.

\begin{table}[!htbp]
\centering
\caption{Uniform versus survival weights: unweighted six-benchmark mean acceptance length $\tau$. Means are reported to three decimal places.}
\label{tab:position-ablation}
\small
\begin{tabular}{llccc}
\toprule
Method & Weights & $T=0$ & $T=0.6$ & $T=1$ \\
\midrule
AdaFlash & Uniform & 6.671 & 6.168 & 5.546 \\
AdaFlash & Survival & 6.691 & 6.219 & 5.568 \\
MDraft & Uniform & 6.568 & 6.260 & 5.836 \\
MDraft & Survival & 6.685 & 6.321 & 5.903 \\
\bottomrule
\end{tabular}
\end{table}

\section{Main Experiment Details}
\label{app:experimental-details}

\textbf{Data and sampled states.} We construct the corpus from the Nemotron-Post-Training-Dataset-v2 dataset. Qwen3-4B generates four independent completions per source prompt with thinking disabled, temperature $0.7$, top-$p=0.8$, top-$k=20$, and an 8,192-token completion cap. We retain 410,448 math sequences and 128,372 code sequences in complete four-completion groups, plus 128,372 sampled chat completion rows; chat prompts need not retain all four attempts. Prompt and completion together are truncated to 1,024 tokens. Filtering sequences too short to provide a full block leaves 665,156 usable sequences. Each epoch samples one anchor uniformly among response positions with room for all 15 subsequent draft tokens. Contexts come from this fixed corpus, while training proposals are generated by the current drafter; teacher continuations are not refreshed during training.

\textbf{Model and optimization.} The initial checkpoint, \texttt{z-lab/Qwen3-4B-DFlash-b16}, has five non-causal layers and a 16-slot layout: one committed anchor and 15 drafted positions ($L=15$ in our notation). Only drafter parameters are updated. We use fp32 trainable parameters with bf16 forward computation, AdamW with learning rate $10^{-5}$, betas $(0.9,0.95)$, epsilon $10^{-8}$, zero weight decay, and gradient-norm clipping at 1.0. The schedule warms up for 3,118 steps and then decays cosinusoidally to zero at step 62,358. Four data-parallel workers each process eight anchors per step, without gradient accumulation; $K=8$ therefore gives 256 sampled blocks but still 32 independent anchors. Completed reference runs use four H100 80 GB GPUs.

\textbf{Objectives and position weights.} AdaFlash samples one greedy block and uses $0.8\mathcal L_{\mathrm{hard}}+0.2\mathcal L_{\mathrm{RKL}}$, with reverse KL evaluated at temperature 1 and its vocabulary-entry contributions capped above at 0.01. Its survival variant applies the same weights to both terms. Our Draft-OPD adaptation uses corpus-sampled anchors, forward KL on the accepted prefix, and reverse KL on the rejected suffix, with equal stream coefficients, separate normalization of nonempty streams, and a rejected-position decay of $0.8^{j-1}$. We reproduce the published system's live rollouts and verification-error replay; however, across our experiments we do not find a detectable gain over using random anchors, but live rollouts significantly increase the computation and time complexity of the method. We hence use random anchors for our implementation of Draft-OPD. MDraft uses $\alpha=0.9$, $K=8$, and temperature-1 proposals; the RKL-only control uses $K=1$ and uniform weights. The reported MDraft runs use RKL as the divergence in Appendix~\ref{app:divergences}; the main table uses survival weights, while Table~\ref{tab:position-ablation} is reserved for the uniform-versus-survival comparison.

\textbf{Checkpoint selection and evaluation.} For the main comparison, we use the final checkpoint at step 62,358 for every fine-tuned method, corresponding to three epochs of training. DFlash is evaluated at the unchanged pretrained checkpoint. Final evaluation covers GSM8K (1,319 prompts), MATH500 (500), AIME25 (30), sanitized MBPP (257), HumanEval (164), and MT-Bench (80, first turn only). Both models decode greedily at $T=0$; at $T=0.6$ and $T=1$, proposal and verifier temperatures match and verification uses exact speculative rejection sampling. Thinking is disabled, the generation cap is 2,048 new tokens, and the evaluation seed is 0. Acceptance length counts committed tokens including the correction/bonus token, lies between 1 and 16, and is averaged within prompts before averaging over each benchmark.

\begin{figure}[!htbp]
\centering
\includegraphics[width=\linewidth]{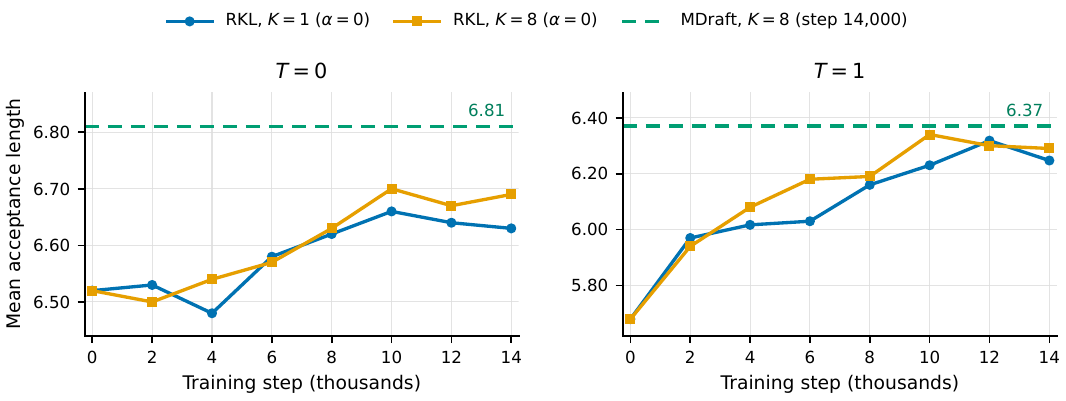}
\caption{Reverse-KL controls with $\alpha=0$ and $K=1$ or $K=8$, evaluated on a GSM8K subset at $T=0$ (left) and $T=1$ (right). Curves include the initial checkpoint and evaluations every 2,000 training steps. Dashed horizontal lines mark MDraft's acceptance length with $K=8$ at step 14,000; they are endpoint references, not MDraft training trajectories.}
\label{fig:rkl-proposal-count}
\end{figure}

\textbf{Multiple-draft reverse-KL control.} We train two controls with $\alpha=0$ and $K\in\{1,8\}$, using Qwen3-4B, the same initial DFlash checkpoint, and the training corpus from the main experiment. At $\alpha=0$, each draft is trained against its own verifier conditional, so increasing $K$ adds verifier evaluations without blending targets. We evaluate mean acceptance length on a GSM8K subset at temperatures $T=0$ and $T=1$. Figure~\ref{fig:rkl-proposal-count} shows similar learning curves for the two controls. At step 14,000, $K=1$ and $K=8$ reach $6.63$ and $6.69$ at $T=0$, and $6.247$ and $6.29$ at $T=1$, respectively. MDraft with $K=8$ reaches $6.81$ and $6.37$ at the same step. Using more drafts with an unchanged RKL target does not recover the gains from target blending.

\FloatBarrier

\subsection{Forward-KL baseline}
\label{app:fkl-baseline}
\begin{figure}[!htbp]
\centering
\includegraphics[width=0.49\linewidth]{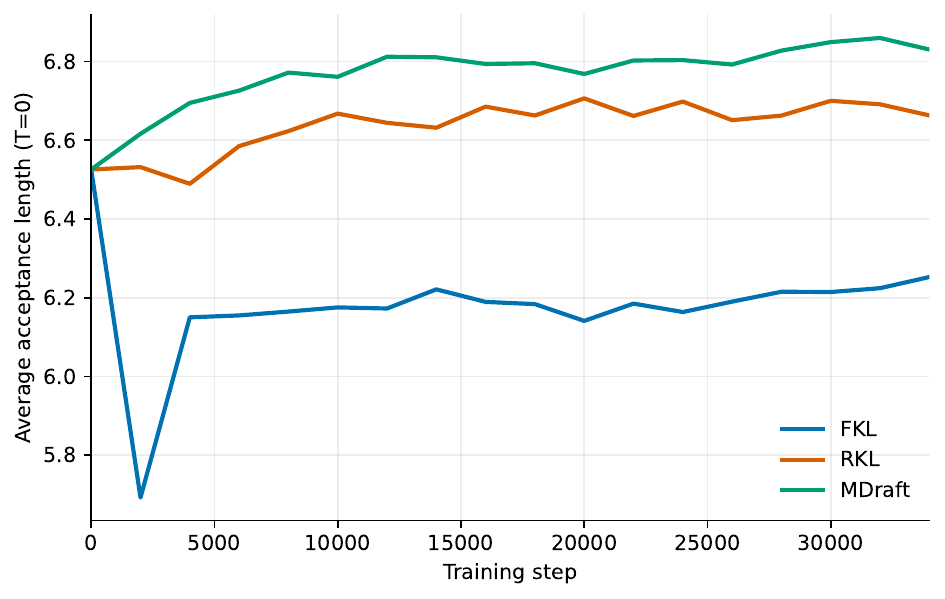}\hfill
\includegraphics[width=0.49\linewidth]{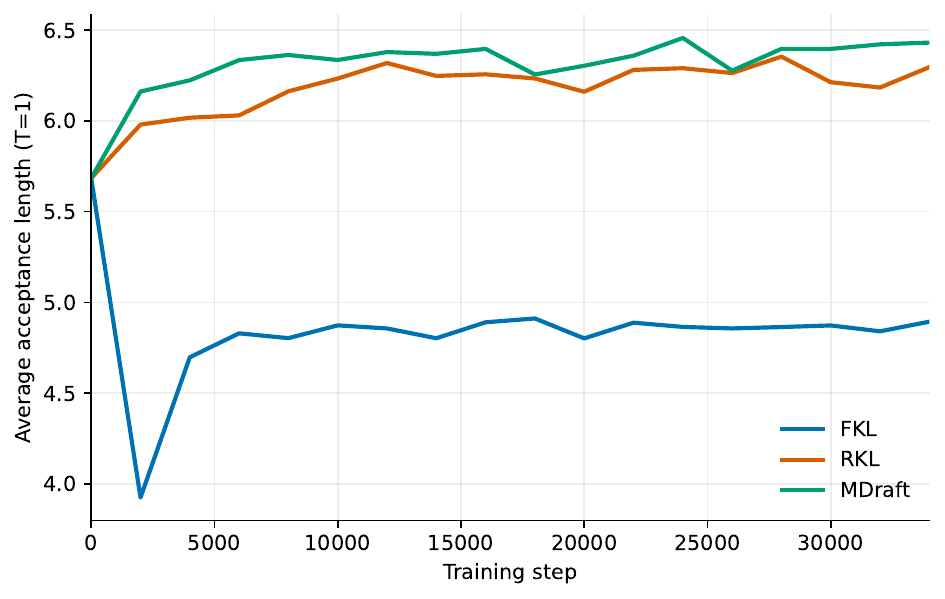}
\caption{Post-training acceptance length on a GSM8K subset at $T=0$ (left) and $T=1$ (right) over 34,000 training steps. Starting from the same pretrained checkpoint, FKL initially reduces acceptance length and only partially recovers, remaining below its initial value. RKL improves over initialization, while MDraft achieves higher acceptance length throughout the subsequent evaluations.}
\label{fig:fkl-posttraining}
\end{figure}

We examine per-prefix forward KL as an additional baseline for post-training. In our experiments, it fails to improve acceptance length when fine-tuning the pretrained DFlash checkpoint. Figure~\ref{fig:fkl-posttraining} shows the post-training dynamics on a GSM8K subset at $T=0$ and $T=1$: FKL remains below the starting checkpoint throughout the measured updates, whereas RKL and MDraft improve acceptance length.

\FloatBarrier

\subsection{Training data for scaling experiments}
\label{app:scaling-data}

For Qwen3-8B, we prepare a training mixture from Nemotron-Post-Training-Dataset-v2 with the same domain proportions as the main experiment: 61.52\% mathematics, 19.24\% code, and 19.24\% chat. Table~\ref{tab:8b-training-data} reports the available source samples and the prepared training mixture. All counts refer to prompt--completion pairs, not unique prompts. We subsample chat pairs with random seed 0 to match the number of retained code pairs, yielding 667,192 training pairs in total.

\begin{table}[!htbp]
\centering
\small
\caption{Composition of the Qwen3-8B training mixture. Counts are prompt--completion pairs.}
\label{tab:8b-training-data}
\begin{tabular}{lrrr}
\toprule
Domain & Available source pairs & Prepared training pairs & Training share \\
\midrule
Math & 410,464 & 410,448 & 61.52\% \\
Code & 135,536 & 128,372 & 19.24\% \\
Chat & 2,182,748 & 128,372 & 19.24\% \\
\midrule
Total & 2,728,748 & 667,192 & 100\% \\
\bottomrule
\end{tabular}
\end{table}

\section{Parallel Drafting and Verification}
\label{app:verification}

Let $c$ denote the context preceding a draft block, $\mathcal V$ the vocabulary, and $L$ the number of newly proposed tokens. We write $p$ for the target model, also called the verifier, and $q$ for the drafter. In the one-step drafting setting, the drafter receives a fully masked block after the context and produces $L$ categorical distributions in a single forward pass:
\begin{equation}
  q(x_{1:L}\mid c)=\prod_{\ell=1}^{L}q_\ell(x_\ell\mid c).
  \label{eq:factorized}
\end{equation}
Tokens are sampled independently across positions given $c$. The drafter's prediction $q_\ell$ for a set token position $\ell$ therefore does not depend on the sampled in-block prefix $x_{<\ell}$. The verifier evaluates
$p_\ell=p(\cdot\mid c,x_{<\ell})$.
All $L$ verifier conditionals can be obtained in one causal forward pass over the proposed block.

Speculative sampling checks proposals from left to right \citep{leviathan2023speculative,chen2023speculative}. If the preceding tokens have been accepted, token $x_\ell$ is accepted with probability
\begin{equation}
  a_\ell(x_{1:\ell})=
  \min\left\{1,\frac{p_\ell(x_\ell)}{q_\ell(x_\ell\mid c)}\right\}.
  \label{eq:accept}
\end{equation}
At the first rejection, a replacement token is sampled from the normalized positive part of $p_\ell-q_\ell$, and the remaining proposals are discarded. If every proposal is accepted, the verifier supplies a bonus token. This correction procedure preserves the verifier's sampling distribution; the sampling-correctness argument below establishes this for the factorized drafter used here.

For a fixed prefix, averaging over the proposed token gives the acceptance probability
\begin{equation}
  a(q,p)=\sum_{y\in\mathcal V}\min\{q(y),p(y)\}
  =1-\mathrm{TV}(q,p),
  \qquad
  \mathrm{TV}(q,p)=\frac12\sum_{y\in\mathcal V}|q(y)-p(y)|.
  \label{eq:tv}
\end{equation}
Block acceptance also depends on whether earlier proposals survive. In particular, the expected number of accepted draft tokens is
\begin{equation}
  \mathbb E[N_{\mathrm{acc}}\mid c]
  =\sum_{j=1}^{L}\mathbb E_{x_{1:L}\sim q(\cdot\mid c)}
  \left[\prod_{\ell=1}^{j}a_\ell(x_{1:\ell})\right].
  \label{eq:acclen}
\end{equation}
This count excludes the correction or bonus token. The product explains why reducing an error early in the block can be more valuable than reducing one near its end.

\textbf{Sampling temperature.} The probabilities in the acceptance rule must be those actually used to sample the target and proposal. For temperature $T>0$, a model with logits $h(y)$ defines probabilities proportional to $\exp(h(y)/T)$; any additional sampling filters must also be reflected in those probabilities. We suppress temperature in the notation when it is clear from context. Under greedy decoding for both models, acceptance reduces to agreement with the verifier's argmax. If only the verifier is greedy and the drafter still samples, the mean acceptance is instead the draft probability assigned to that argmax.

\phantomsection\label{app:sampling}
\textbf{Sampling correctness.} We specialize the standard speculative-sampling argument \citep{leviathan2023speculative,chen2023speculative} to the factorized proposal in \eqref{eq:factorized}. Fix the context $c$ and an accepted prefix $x_{<\ell}$. Abbreviate $p(y)=p(y\mid c,x_{<\ell})$ and $q(y)=q_\ell(y\mid c)$. Because draft positions are sampled independently, conditioning on the acceptance of earlier positions leaves the current proposal distributed as $q$.

For $X\sim q$, accept with probability $a(X)=\min\{1,p(X)/q(X)\}$, evaluated only on the support of $q$. Let
\begin{equation}
 A=\sum_y\min\{p(y),q(y)\},\qquad
 p_{\mathrm{res}}(y)=\frac{[p(y)-q(y)]_+}{1-A}\quad\text{if }A<1,
\end{equation}
where $[u]_+=\max\{u,0\}$. The residual is normalized because $\sum_y[p(y)-q(y)]_+=1-A$. The joint probability of proposing and accepting token $y$ is $q(y)a(y)=\min\{p(y),q(y)\}$. Therefore the emitted token $Y$, including correction after rejection, satisfies
\begin{align}
 \Pr(Y=y)
 &=\Pr(X=y,\text{accept})+\Pr(\text{reject})p_{\mathrm{res}}(y)\nonumber\\
 &=\min\{p(y),q(y)\}+[p(y)-q(y)]_+=p(y).
\end{align}
If $A=1$, then $p=q$ and no correction is needed. Applying this identity successively to accepted positions, discarding unused proposals after a rejection, and sampling the bonus token from the verifier proves preservation of its autoregressive distribution. The argument uses the actual evaluation distributions, including any temperature or sampling filters, and does not depend on the training objective.

\end{document}

%% file: math_commands.tex
\usepackage{amsmath,amsfonts,bm}

\def\eqref#1{equation~\ref{#1}}

\def\1{\bm{1}}

\DeclareMathAlphabet{\mathsfit}{\encodingdefault}{\sfdefault}{m}{sl}
\SetMathAlphabet{\mathsfit}{bold}{\encodingdefault}{\sfdefault}{bx}{n}